\documentclass[journal]{IEEEtran}
\usepackage[utf8]{inputenc} 
\usepackage[T1]{fontenc}    
\usepackage{hyperref}       
\usepackage{url}            
\usepackage{booktabs}       
\usepackage{amsfonts}       
\usepackage{nicefrac}       
\usepackage{microtype}      
\usepackage{xcolor}         

\usepackage{color,amsmath,amstext,amsfonts,amssymb, mathrsfs,mathtools}
\usepackage{graphicx,algorithmic,algorithm}
\usepackage{tikz}
\usepackage{setspace}
\usepackage{array} 
\usepackage{booktabs}  
\usepackage{bbm}
\usepackage{wasysym}
\usepackage{textcomp}
\usepackage[sort,compress]{cite}
\usepackage{mathtools}
\usepackage{empheq}

\newtheorem{lemma}{Lemma}
\newtheorem{assumption}{Assumption}

\newtheorem{theorem}{Theorem}

\let\oldnl\nl
\newcommand{\nonl}{\renewcommand{\nl}{\let\nl\oldnl}}

\newcounter{l1}
\newcounter{l2}
\newcounter{l3}
\newcommand{\bdotlist}{\begin{list}{$\bullet$}{}}
\newcommand{\bboxlist}{\begin{list}{$\Box$}{}}
\newcommand{\bbboxlist}{\begin{list}{\raisebox{.005in}{{\tiny $\blacksquare$ \ \ }}}{}}
\newcommand{\bdashlist}{\begin{list}{$-$}{} }
\newcommand{\blist}{\begin{list}{}{} }
\newcommand{\barablist}{\begin{list}{\arabic{l1}}{\usecounter{l1}}}
\newcommand{\balphlist}{\begin{list}{(\alph{l2})}{\usecounter{l2}}}
\newcommand{\bAlphlist}{\begin{list}{\Alph{l2}.}{\usecounter{l2}}}
\newcommand{\bdiamlist}{\begin{list}{$\diamond$}{}}
\newcommand{\bromalist}{\begin{list}{(\roman{l3})}{\usecounter{l3}}}

\newcommand{\norm}[1]{\left\lVert#1\right\rVert}

\newcommand{\beq}{\begin{equation}}
\newcommand{\eeq}{\end{equation}}

\newcommand{\tn}{\textnormal}

\allowdisplaybreaks

\title{Online Convex Optimization with Long Term Constraints for Predictable Sequences}

\author{Deepan Muthirayan,~\IEEEmembership{Member,~IEEE}, Jianjun Yuan, and Pramod P. Khargonekar ~\IEEEmembership{Fellow,~IEEE}
\thanks{This work is supported in part by the National Science Foundation under Grant ECCS-1839429.
D. Muthirayan and P. P. Khargonekar are with the Department of Electrical Engineering and Computer Sciences, University of California Irvine, Irvine, CA (emails: deepan.m@uci.edu, pramod.khargonekar@uci.edu). Jianjun Yuan is with the Expedia Group (email: yuanx270@umn.edu).}
}

\begin{document}

\maketitle 
\thispagestyle{empty}
\pagestyle{empty}

\begin{abstract}
In this paper, we investigate the framework of Online Convex Optimization (OCO) for online learning. OCO offers a very powerful online learning framework for many applications. In this context, we study a specific framework of OCO called {\it OCO with long term constraints}. Long term constraints are introduced typically as an alternative to reduce the complexity of the projection at every update step in online optimization. While many algorithmic advances have been made towards online optimization with long term constraints, these algorithms typically assume that the sequence of cost functions over a certain $T$ finite steps that determine the cost to the online learner are adversarially generated. In many circumstances, the sequence of cost functions may not be unrelated, and thus predictable from those observed till a point of time. In this paper, we study the setting where the sequences are predictable. We present a novel online optimization algorithm for online optimization with long term constraints that can leverage such predictability. We show that, with a predictor that can supply the gradient information of the next function in the sequence,
our algorithm can achieve an overall regret and constraint violation rate that is strictly less than the rate that is achievable without prediction. 
\end{abstract}


\section{Introduction}

Learning can be key to intelligent decision making especially under circumstances where the underlying environment is uncertain or unknown. Many practical circumstances are scenarios where the learner does not have access to large history of data and has to learn from data gathered online, i.e., while operating in the environment. Such a learning problem is broadly termed as {\it online learning}. Online media platforms like Netflix, Spotify etc. \cite{hazan2016introduction} leverage online learning techniques in their algorithms. On such platforms, the central problem is to decide what to recommend to the users. Given that the platforms do not have access to prior user data, the platforms will have to learn the users' preferences online. It is intuitively clear that online learning is relevant to dynamical systems and control. It is noteworthy that the application of online learning to control has received significant interest in the recent years \cite{hazan2020nonstochastic, muthirayan2021online}. There are several other practical applications of online learning like portfolio selection, 
online display advertising \cite{goldfarb2011online}, learning from experts \cite{hazan2016introduction} etc.

In a typical online learning setting, the online learner can make its decision utilizing the history of data it has seen so far. For example, in the context of recommendation systems, it can decide which movies or shows to recommend to a user given the history of user's response to the recommendations. Since the decisions are made online, the decision typically incurs a cost. In the above example, the cost is a function of user's response to the recommendations. This is because, typically, the users' positive response to the recommendations translates into profits.  The  incurred cost serves as feedback to the online learner, which it can then use to improve its decision in the next iteration. The cost feedback can be broadly classified as: (i) full information feedback and (ii) bandit feedback. Full information feedback is the case where the feedback can be used to infer the full cost function for the time step and bandit feedback is the case where the feedback is just the incurred cost, which cannot be directly used to infer the cost function. The effectiveness of the online learner lies in how effectively it is able to learn with the incremental feedback it receives at every step along the way.

Online Convex Optimization (OCO) is a very widely studied online decision framework for such online learning scenarios. 
The power and wide reach of OCO lies in the fact that it is applicable to many of the online learning scenarios mentioned above \cite{hazan2016introduction}. We study a specific extension of the standard OCO framework, the scenario when the underlying sequence of costs are predictable. The traditional OCO framework assumes that the sequence of cost functions is arbitrary and has no patterns. In several practical scenarios, this can be a very permissive assumption. It is conceivable that the predictability inherent in the sequence of costs can be leveraged to improve the sequence of decisions made online. Not surprisingly, the idea of predictions (assuming the predictions are supplied by an oracle) have been used to improve OCO algorithms (see \cite{ rakhlin2013online, bhaskara2020online}). 

In this work, we specifically consider the development of OCO algorithms that can leverage predictions for an extension of the OCO framework called {\it OCO with long term constraints}. OCO with long term constraints was introduced as a framework to reduce the computational burden that arises as a result of the projection step at every update step. We elaborate the framework of OCO with long term constraints and its relevance in the next section. We then review a standard algorithm for OCO with long term constraints and then introduce our algorithm that can leverage the predictions offered by an oracle.

\subsection{Contribution}

We present a novel online optimization algorithm for OCO with long term constraints that can leverage predictions. We show that, with a predictor that can supply the gradient information with error $\mathcal{O}(T^{-a/2})$, $a \in (0,1)$ (which covers all scenarios), 
our algorithm can achieve a $\mathcal{O}(T^{(1-a)/2})$ regret and a $\mathcal{O}(T^{1/4+a/4})$ constraint violation, i.e., an overall rate that is strictly less than $\mathcal{O}(\sqrt{T})$, the achievable rate without prediction. 
{\it Our main contribution is the algorithmic methodology itself. Our proof technique draws on the techniques from the literature in online optimization. Specifically, our analysis technique builds on the most recent approach \cite{yi2021regret}, but can incorporate time varying constraints unlike \cite{yi2021regret}}. We state that it is an open problem to simultaneously reduce both the regret and constraint violation from what is achievable without prediction.


\subsection{Related Works}

The problem of regret minimization with long term constraints has been extensively studied in \cite{mahdavi2012trading, jenatton2016adaptive, yu2017online, yuan2018online, yu2016low, yi2021regret}. 
The literature so far has established that a regret of $\mathcal{O}(\sqrt{T})$ and a cumulative constraint violation of $T^{1/4}$ can be achieved. But none of these works study the OCO problem with predictability in the cost function sequence.

Many prior works \cite{hazan2007online, dekel2017online} study the problem of optimization with some notion of predictability. 
In \cite{bhaskara2020online}, the authors study the case where the prediction may not be helpful at all times. The authors present an algorithm that smoothly interpolates between the two extreme cases, i.e., when the hints are good at all times and when the hints are bad at all times. In \cite{rakhlin2013online}, the authors present an alternate approach to the problem of optimizing with predictions (or hints). They derive a regret bound in terms of the cumulative error of the predictions (or hints). But none of these works consider the constraints formulated as long term constraints.

\subsection{Notations}

We use $\norm{\cdot}$ for the norm. Additionally, $\norm{\cdot}_p$ denotes the $p$-norm, for eg., $\norm{\cdot}_2$ denotes the two norm and $\norm{\cdot}_1$ denotes the one norm. We denote the transpose of a vector $x$ by $x^\top$. We denote the projection on to a set $\mathcal{K}$ by $\tn{Proj}_{\mathcal{K}}(\cdot)$. We denote the sequence $\{x_1, \dots, x_T\}$ compactly by $x_{1:T}$. We denote the Bregman divergence of a strongly convex function $\mathcal{R}$ by $D_\mathcal{R}(\cdot, \cdot)$. $\langle x,y \rangle$ denotes the standard inner product of two vectors $x,y$. The function $\{\cdot\}_{+}$ denotes the non-negative component of the input.

\section{Problem Formulation}
We consider the problem of online optimization with long term constraints and with predictable sequences. We adopt the standard notation. Every iteration or step in online optimization is typically indexed by time $t$ and is associated with a convex cost function $f_t(\cdot)$. At every step $t$, the decision maker makes a decision, $x_t \in \mathcal{X} \subseteq \mathbb{R}^p$ 
and incurs the cost $f_t(x_t)$. Typically, the decision has to lie within a set $\mathcal{K} \subseteq \mathcal{X}$. The challenge is that the cost functions are unknown apriori and can be chosen by an adversary. Thus, this problem can be seen as a game between the decision maker and an adversary which can choose the cost functions arbitrarily. The decision maker upon its decision $x_t$ receives a feedback on its decision. The feedback can be {\it full information}, in which case the full cost function $f_t(\cdot)$ or the gradient of the cost $\partial f_t(x_t)$ is revealed along with the realized cost $f_t(x_t)$, or the feedback can be {\it bandit}, in which case no further information is revealed apart from the realized cost $f_t(x_t)$. Thus, the decision maker can use this information to improve the decision at the next time step. The goal is to compute a sequence of decisions $x_1, x_2, \ldots$ with the feedback received at every step along the way to minimize the regret
\beq
R_T = \sum_{t=1}^T f_t(x_t) - \min_{x \in \mathcal{K}} \sum_{t=1}^T f_t(x), \nonumber 
\eeq 
where $\min_{x \in \mathcal{K}} \sum_{t=1}^T f_t(x)$ is the full cost of the optimal decision. 
To give an example, in recommendation systems, the optimal decision is the optimal recommendation to all the users and the regret as defined above captures how well the system is able to learn and improve its recommendation over the course of time. Thus, the regret is a well defined measure of how well the decision maker is able to learn online. 

For the above problem, it is well known that the simple and projection based online gradient algorithm,
\beq x_{t+1} = \text{Proj}_\mathcal{K}(x_t - \eta \partial f_t(x_t)), \nonumber \eeq 
where $\text{Proj}_\mathcal{K}(\cdot)$ denotes the projection onto the set $\mathcal{K}$, when $\mathcal{K}$ is closed and convex, achieves $\mathcal{O}(\sqrt{T})$ when the sub-gradients are bounded \cite{zinkevich2003online}, where $\mathcal{O}(\sqrt{T})$ has been shown to be tight. 

Despite the simplicity of the gradient algorithm, the projection step of the algorithm can be computed easily or in closed form only for simple sets such as a ball or a box. In general, the projection step can be computationally expensive for a general convex set of the form $\mathcal{K}: \{x: g(x) \leq 0\}$, where $g(x) = [g^1(x), g^2(x),...,g^m(x)]^\top$, with $g^i : \mathbb{R}^p \rightarrow \mathbb{R}$ being a convex function. An alternate framework was introduced in \cite{mahdavi2012trading} to simplify the approach. Here, instead of requiring that $g(x) \leq 0$ in each step, the constraint is required to be satisfied only in the long run. Thus, this formulation is popularly referred to as {\it OCO with long term constraints}. Therefore, the goal here is, in addition to minimizing the regret, to minimize the cumulative constraint violation given by
\beq C_T = \sum_{t=1}^T \sum_j \{g^j_t(x_t)\}_+. \label{eq:constraintviolation} \eeq 
In particular, the objective is to achieve sub-linear regret and sub-linear cumulative constraint violation. This ensures that, in the long run, the decision satisfies the constraints and the regret only grows sub-linearly. Since these constraints are typically known apriori, we assume that the function sequence $g_t$s are known.

We make the following standard assumption on the cost function $f_t$ and the constraint function $g_t$.
\begin{assumption}
(i) The set $\mathcal{X}$ is a closed and a convex set. (ii) The constraint function $g_t$ is convex for all $t$. (iii) The cost function $f_t$ is linear, i.e, $f_t(x) = \langle c_t, x_t \rangle $. (iv) The sub-gradient $\partial f_t(\cdot) = c_t \leq G, ~ \partial g_t \leq G ~ \forall ~ x \in \mathcal{X}$, where $G$ is a constant. (v) The functions $f_t(x) \leq F, g_t(x) \leq F, ~ \forall ~ x \in \mathcal{X}$. (vi) The decision maker receives full information feedback.
\label{as:system}
\end{assumption}

The assumption that the cost functions are linear is standard in the OCO setting with predictable sequences; see \cite{rakhlin2013online, bhaskara2020online}. The other assumptions are also standard in the online optimization literature \cite{yi2021regret}. As a start, we focus here on the full information feedback setting and we believe our algorithms can be extended to the bandit feedback setting just as in the setting without prediction.


As motivated earlier, the assumption that the sequence of cost functions are arbitrary can be harsher. 
The idea here is that the sequence of costs might not be totally arbitrary and adversarial but have a relation. In such circumstances, it is possible that the decision maker can predict the next cost from the past information, and thus improve its performance. Therefore, we consider the setting where the sequence of cost functions are predictable and there is an oracle that supplies a prediction $h_t$ of the gradient $c_t$ before time $t$, very much in the spirit of \cite{rakhlin2013online, bhaskara2020online}.  Our goal is to develop an algorithm for the OCO setting with long term constraints that can leverage such predictions to improve the regret while keeping the constraint violation controlled.

\section{Review: OCO Algorithm}

Here, we review the most recent approach to online optimization with long term constraints formulation \cite{yi2021regret}. We present the algorithm and its properties for the case where the constraint functions are time invariant, i.e., $g_t = g$. A standard algorithm for OCO with long term constraints is shown in Algorithm \ref{alg:online-alg-standard}.  

\begin{algorithm}[]
\begin{algorithmic}[1]
\STATE \textbf{Initialize}: $x_1 \in \mathcal{X}, q_0 = 0, \gamma_1$
\FOR{$t = 1,\ldots,T$}
\STATE Apply $x_t$; suffer cost $f_t(x_t)$; suffer constraint violation $[g(x_t)]_+$ 
\STATE Observe $c_t$
\STATE Set $\eta_{t}$ and $\gamma_{t+1}$
\STATE Update
\begin{align}
 q_{t} &  = q_{t-1} + \gamma_{t} [g(x_{t})]_{+}, ~ \hat{q}_t  = q_{t} + \gamma_{t} [g(x_{t})]_{+}, \nonumber \\
x_{t+1} & = \arg\min_{x \in \mathcal{X}} \eta_{t} \langle x , c_t \rangle \nonumber \\
& + \eta_{t} \gamma_{t+1} \langle \hat{q}_t , [g(x)]_{+}\rangle  + \norm{x - x_t}^2_2. \nonumber 
\end{align} 
\ENDFOR
\end{algorithmic}
\caption{Online Optimization Algorithm without Prediction \cite[Algorithm 1]{yi2021regret}}
\label{alg:online-alg-standard}
\end{algorithm}

The key difference in this algorithm compared to a standard OCO algorithm is the optimization step carried out to compute $x_{t+1}$. A standard online algorithm just updates the decision along the direction of the gradient of the current cost with an appropriate step size. In this case though, the decision is updated along a direction that is a combination of the gradient of the cost function and a direction that minimizes the cumulative constraint violation. This additional direction compared to the standard algorithm is contributed by the second term inside the optimization carried out to update $x_t$ to $x_{t+1}$. This design achieves the desired properties in the long run, i.e., achieves sub-linear constraint violation and sub-linear regret. We summarize the properties of the algorithm below.

\begin{theorem}[Theorem 1. \cite{yi2021regret}]
Suppose Assumption \ref{as:system} holds. Suppose $\eta_t = 1/T^c, \gamma_t = \frac{1}{G\sqrt{2\eta}}$, where $c \in (0,1)$. Then, Algorithm \ref{alg:online-alg-standard} achieves
\beq
R_T = \mathcal{O}(T^{\max\{1-c,c\}}), ~ C_T \leq T^{1/2-c/2}. \nonumber 
\eeq 
\end{theorem}

The proof of this Thoerem is available in \cite{yi2021regret}. 

\section{OCO Algorithm with Prediction}

In this scenario, the following information is available to decide $x_t$ at $t$: the prediction $h_t$ for the gradient of the cost function at $t$, in addition to the gradient of the cost functions at all previous time steps. With prediction, the algorithm can afford to anticipate and therefore make better decisions. The online algorithm we propose leverages the prediction $h_t$ to make its decision $x_t$ at time $t$. The complete algorithm is outlined in Algorithm \ref{alg:online-alg}.  

\begin{algorithm}[]
\begin{algorithmic}[1]
\STATE \textbf{Input}: $\mathcal{R}$ $1-$strongly convex function w.r.t $\norm{\cdot}_2$
\STATE \textbf{Initialize}: $x_1 = z_1 = \arg\min_{z \in \mathcal{X}} \mathcal{R}(z), q_0 = 0$
\FOR{$t = 1,\ldots,T$}
\STATE Apply $x_t$; suffer cost $f_t(x_t)$; suffer constraint violation $[g_t(x_t)]_+$ 
\STATE Observe $c_t$; observe the hint $h_{t+1}$
\STATE Set $\eta_{t+1}$ and $\gamma_{t+1}$
\STATE Update
\begin{align}
 q_{t}&  = q_{t-1} + \gamma_{t} [g_{t}(x_{t})]_{+}\nonumber \\  z_{t+1} & = \arg\min_{z \in \mathcal{X}} \eta_{t+1} \langle z , c_t \rangle \nonumber \\
& + \eta_{t+1} \gamma_{t}  \langle \hat{q}_{t-1} , [g_{t}(z)]_{+}\rangle  + D_\mathcal{R}(z,z_t). \nonumber 
\end{align} 
\STATE Update
\begin{align}
\hat{q}_t & = q_{t} + \gamma_{t+1} [g_{t+1}(z_{t+1})]_{+}\nonumber \\
x_{t+1} & = \arg\min_{x \in \mathcal{X}} \eta_{t+1} \langle x , h_{t+1} \rangle \nonumber \\
& + \eta_{t+1} \gamma_{t+1} \langle \hat{q}_t , [g_{t+1}(x)]_{+}\rangle  + D_\mathcal{R}(x,z_{t+1}). \nonumber 
\end{align}
\ENDFOR
\end{algorithmic}
\caption{Online Optimization Algorithm with Prediction}
\label{alg:online-alg}
\end{algorithm}

Our algorithm has two update steps. In each update step a constraint violation cost is included, which is needed to simultaneously minimize the constraint violation alongside the regret. The intuition for this inclusion is the same as Algorithm \ref{alg:online-alg-standard}. The two step decision update is similar to the two step update in the standard online optimization with prediction. The first step computes a decision like in the regular online optimization with long term constraints. 
The second update step updates the decision further along the prediction $h_t$ as in the standard online update with prediction. The difference here is that the optimization objective for this step like in the first step includes an additional constraint violation cost. The additional constraint violation cost included in the second update step has the same form as the constraint violation cost in standard online update with long term constraints \cite{yi2021regret}. We note that there are differences in how $\hat{q}_t$s is defined and the constraint violation cost in each update step. 

Next, we present the regret and the cumulative constraint violation our algorithm can achieve with prediction. 
\begin{theorem}
Suppose Assumption \ref{as:system} holds. Suppose $\eta_t = \eta = 1/T^{c}, \gamma_t = \gamma = \frac{1}{G\sqrt{\eta}}$, where $c \in (0,1)$. Then, under Algorithm \ref{alg:online-alg}, when $\norm{c_t  - h_t} = \mathcal{O}\left(T^{-a/2}\right)$,
    \beq R_T \leq \mathcal{O}(T^{\max\{1-a-c,c\}}), ~ C_T \leq T^{1/2-c/2}. \nonumber \eeq 
\label{thm:online-alg}
\end{theorem}

{\bf Discussion}: First, we note that, we recover the state-of-the-art result when the prediction does not provide any valuable information, that is, when $a = 0$: $\mathcal{O}(\sqrt{T})$ for regret and $\mathcal{O}(T^{1/4})$ for cumulative constraint violation. We can recover the same result for any $a$ by setting $c=1/2$. But, we can do better and reduce the regret further.
When the prediction is the worst, since $c_t \leq G$, $\norm{c_t - h_t} = \mathcal{O}(G) = \mathcal{O}(1)$. 
We can set $a$ more tighter when the prediction error is small. For example, when $\norm{c_t - h_t} = \mathcal{O}\left(T^{-b}\right)$, where $b \in [0,1/2)$, we can set $a = 2b$, and when $\norm{c_t - h_t} = \mathcal{O}\left(T^{-b}\right)$, where $b \geq 1/2$, we can still set $a < 1$. Across these scenarios, setting $c = (1-a)/2$ gives us $\mathcal{O}(T^{(1-a)/2})$ regret and $\mathcal{O}(T^{1/4+a/4})$ constraint violation, 
which gives us a much better regret but with an increase in constraint violation. The upside is that the constraint violation rate is still within $o(\sqrt{T})$, which is strictly less than the $\mathcal{O}(\sqrt{T})$ overall rate that is achievable in any form OCO without prediction. Here, the designer can choose to set $a$, for eg., when $b \geq 1/2$, depending on how much less of constraint violation, given by $\mathcal{O}(T^{1/4+a/4})$, is desirable compared to $\mathcal{O}(\sqrt{T})$.
We state that it is an open problem to simultaneously reduce regret and constraint violation. We suspect this could be a result of certain theoretical bottlenecks prevalent in all OCO analysis. We point to this bottleneck in the next section.

\section{Main Results}
We break the analysis in to two main steps. In the first lemma, we derive two key inequalities relating the variables of the problem. We then use this intermediate lemma to prove our theorem. 
\begin{lemma}
Suppose Assumption \ref{as:system} holds. Then, for any $x^{*} \in \mathcal{X}$, under Algorithm \ref{alg:online-alg}
\begin{align}
& \sum_{t=1}^T \gamma_t \langle q_{t-1}, [g_t(x_t)]_{+} \rangle + R_T \leq \sum_{t=1}^T \frac{\eta_{t}}{2}\norm{c_t - h_t}^2 \nonumber \\
& + \sum_{t=1}^T \frac{1}{\eta_t}\left(D_\mathcal{R}(z_t,x^{*}) - D_\mathcal{R}(x^{*},z_{t+1})\right) - \sum_{t=1}^T \frac{1}{\eta_t} D_\mathcal{R}(x_t,z_t) \nonumber \\
& + \sum_{t=1}^T\gamma_t \langle \hat{q}_{t-1}, [g_t(x^{*})]_{+} \rangle - \sum_{t=1}^T\gamma^2_t  \langle [g_t(z_t)]_{+},[g_t(x^{*})]_{+} \rangle,  \nonumber \\
& \frac{1}{2}\norm{q_T}^2 \leq \sum_{t=1}^T \frac{\eta_{t}}{2}\norm{c_t - h_t}^2 + \sum_{t=1}^T \gamma_t \langle \hat{q}_{t-1}, [g_t(x^{*})]_{+} \rangle -R_T \nonumber \\
&  + \sum_{t=1}^T\frac{1}{\eta_t}\left(D_\mathcal{R}(z_t,x^{*}) - D_\mathcal{R}(x^{*},z_{t+1})\right) \nonumber \\
& + \sum_{t=1}^T\left(\gamma^2_t G_t^2 - \frac{1}{\eta_t}\right) D_\mathcal{R}(x_t,z_t). \nonumber
\end{align}
\label{lem:intermediate}
\end{lemma}

Please see Appendix \ref{sec:lemma} for the proof. In the next theorem, we use the above lemma to derive a bound on the regret and cumulative constraint violation in terms of the cumulative error in the prediction.

\begin{theorem}
Suppose Assumption \ref{as:system} holds, $\eta_t = \eta, \gamma_t = \gamma = \frac{1}{G\sqrt{\eta}}$, then for any $x^{*} \in \{x: g_t(x) \leq 0, \forall t\}$, Algorithm \ref{alg:online-alg} achieves
\begin{align}
& R_T \leq \frac{\eta}{2} \sum_{t=1}^T \norm{c_t - h_t}^2 + \frac{1}{\eta} D_\mathcal{R}(z_1,x^{*}), \nonumber \\
& C_T \leq  \frac{1}{\gamma}\sqrt{\frac{\eta}{2} \sum_{t=1}^T \norm{c_t - h_t}^2 + \frac{1}{\eta} D_\mathcal{R}(z_1,x^{*}) + FT}. \nonumber
\end{align}
\label{thm:linear-case}
\end{theorem}

Please see Appendix \ref{sec:thm} for the proof. It is easy to see that when the prediction error is not very small, i.e., $\norm{c_t - h_t}_2 = \mathcal{O}(1)$, the regret upper bound reduces to the standard bound of the form $\mathcal{O}(\eta T) + \mathcal{O}(1/\eta)$. Here, we can set the step rate $\eta = \mathcal{O}(T^{-1/2})$ and recover the rate that is achievable without prediction. Similarly, for this case, we can recover the rate $\mathcal{O}(T^{1/4})$ for cumulative constraint violation, and thereby recover the scaling that is achievable without prediction. 

When the prediction error is small, lets consider the case where $\norm{c_t  - h_t} \leq 1/\sqrt{T^a}$. With $\eta_t$ set as $\eta_t = \eta = T^{-c}$, where $c$ is some constant and $c \in (0,1)$, it follows that 
\begin{align}
& \frac{\eta}{2} \sum_{t=1}^T \norm{c_t - h_t}^2 \leq \mathcal{O}(T^{1-a-c}), ~ \frac{1}{\eta} D_\mathcal{R}(z_1,x^{*}) \leq \mathcal{O}(T^{c}) \nonumber \\
& \Rightarrow R_T \leq \mathcal{O}(T^{\max\{1-a-c,c\}}). \nonumber 
\end{align}

Similarly, we have that
\begin{align}
& \frac{\eta}{2} \sum_{t=1}^T \norm{c_t - h_t}^2 + \frac{1}{\eta} D_\mathcal{R}(z_1,x^{*}) \leq \mathcal{O}(T^{\max\{1-a-c,c\}}) \nonumber \\
& \frac{1}{\gamma} = G T^{-c/2}, ~ \Rightarrow C_T \leq \mathcal{O}(T^{\frac{\max\{1-a-c,c,1\}}{2}}) T^{-c/2} \nonumber \\
& =  \mathcal{O}(T^{1/2-c/2}). \nonumber 
\end{align}

This completes the analysis and proof of Theorem \ref{thm:online-alg}. The bottleneck arises from the ``1'' term in $\max\{1-a-c,c,1\}$. This arises from the $FT$ term in the upper bound to $C_T$, which ends up dominating the other two terms. The term $FT$ arises from having to bound $-R_T$. To the best of our knowledge, there are no known better bounds than $FT$ and this is the technical bottleneck.

\section{Conclusion}

In this work, we consider the development of an algorithm that can leverage the predictability in the sequence of cost functions in online optimization. We specifically consider the online optimization setting with the constraints formulated as long term constraints. 
For this setting, we present a novel online optimization algorithm that can leverage predictions. We show that our algorithm can achieve an overall regret and cumulative constraint rate that is strictly less than $\mathcal{O}(\sqrt{T})$, the rate that is achievable without prediction. We also present some open challenges. 

\bibliographystyle{IEEEtran}
\bibliography{Refs}

\section*{Appendix A: Proof of Lemma \ref{lem:intermediate}}
\label{sec:lemma}

For any $x^{*} \in \mathcal{X}$,
\begin{align}
& R_T = \langle x_t - x^{*}, c_t \rangle \leq \langle x_t - z_{t+1}, c_t - h_t \rangle \nonumber\\
& + \langle x_t - z_{t+1}, h_t + \gamma_t \partial [g_t(x_t)]_{+}\hat{q}_{t-1} \rangle \nonumber \\
& + \langle z_{t+1} - x_t , \gamma_t \partial [g_t(x_t)]_{+}\hat{q}_{t-1} \rangle \nonumber \\
& + \langle z_{t+1} - x^{*}, c_t + \gamma_t \partial [g_t(z_{t+1})]_{+}\hat{q}_{t-1} \rangle  \nonumber \\
& + \langle x^{*} - z_{t+1}, \gamma_t \partial [g_t(z_{t+1})]_{+}\hat{q}_{t-1} \rangle. \nonumber 
\end{align} 

Applying Cauchy-Schwarz to the first term we get
\begin{align}
& \langle x_t - z_{t+1}, c_t - h_t \rangle \leq \norm{ x_t - z_{t+1}}\norm{c_t - h_t} \nonumber \\
& \leq \frac{1}{2\eta_{t}} \norm{x_t - z_{t+1}}^2 + \frac{\eta_{t}}{2}\norm{c_t - h_t}^2. \label{eq:pf1-eq1} 
\end{align}

Now, for any convex function $h(\cdot)$, any update of the form $a^{*} = \arg\min_{a \in \mathcal{A}} h(a) + D_\mathcal{R}(a, b)$ (see \cite{yi2021regret}), for any $d \in \mathcal{A}$, satisfies 
\beq \langle a^{*} - d , \partial h(a^{*}) \rangle \leq D_\mathcal{R}(d,b) - D_\mathcal{R}(d,a^{*}) - D_\mathcal{R}(a^{*},b). \nonumber \eeq 

Applying this to the second term with $h(a) = \langle a, h_t\rangle + \gamma_t \langle \hat{q}_{t-1}, [g_t(a)]_+ \rangle $, and using the second update step of Algorithm\ref{alg:online-alg}, we get
\begin{align} 
& \langle x_t - z_{t+1}, h_t + \gamma_t \partial [g_t(x_t)]_{+}\hat{q}_{t-1} \rangle \nonumber \\
& \leq \frac{1}{\eta_t}\left(D_\mathcal{R}(z_t,z_{t+1}) - D_\mathcal{R}(z_{t+1},x_t) - D_\mathcal{R}(x_t,z_t)\right). \label{eq:pf1-eq2}
\end{align}

Applying the same formula to the fourth term with $h(a) = \langle a, c_t\rangle + \gamma_t \langle \partial [g_t(z_{t+1})]_{+}\hat{q}_{t-1}, a\rangle $, and using the first update step of Algorithm\ref{alg:online-alg}, we get
\begin{align} 
& \langle z_{t+1} - x^{*}, c_t + \gamma_t \partial [g_t(z_{t+1})]_{+}\hat{q}_{t-1} \rangle \nonumber \\
& \leq \frac{1}{\eta_t}\left(D_\mathcal{R}(z_t,x^{*}) - D_\mathcal{R}(x^{*},z_{t+1}) - D_\mathcal{R}(z_{t+1},z_t)\right). 
\label{eq:pf1-eq3} 
\end{align}

Then, using Equations \eqref{eq:pf1-eq1}, \eqref{eq:pf1-eq2}, and \eqref{eq:pf1-eq3}, we get
\begin{align}
& \langle x_t - x^{*}, c_t \rangle \leq \frac{1}{2\eta_{t}} \norm{x_t - z_{t+1}}^2 + \frac{\eta_{t}}{2}\norm{c_t - h_t}^2 \nonumber \\
& + \frac{1}{\eta_t}\left(D_\mathcal{R}(z_t,z_{t+1}) - D_\mathcal{R}(z_{t+1},x_t) - D_\mathcal{R}(x_t,z_t)\right) \nonumber \\
& + \frac{1}{\eta_t}\left(D_\mathcal{R}(z_t,x^{*}) - D_\mathcal{R}(x^{*},z_{t+1}) - D_\mathcal{R}(z_{t+1},z_t)\right) \nonumber \\
& + \langle z_{t+1} - x_t , \gamma_t \partial [g_t(x_t)]_{+}\hat{q}_{t-1} \rangle \nonumber \\
& + \langle x^{*} - z_{t+1}, \gamma_t \partial [g_t(z_{t+1})]_{+}\hat{q}_{t-1} \rangle.  \nonumber
\end{align} 

Since $D_\mathcal{R}(z_{t+1},x_t) \geq \frac{1}{2} \norm{x_t - z_{t+1}}^2$ by definition, we get
\begin{align}
& \langle x_t - x^{*}, c_t \rangle \leq \frac{\eta_{t}}{2}\norm{c_t - h_t}^2 \nonumber \\
& + \frac{1}{\eta_t}\left(D_\mathcal{R}(z_t,x^{*}) - D_\mathcal{R}(x^{*},z_{t+1})- D_\mathcal{R}(x_t,z_t)\right) \nonumber \\
& + \langle z_{t+1} - x_t , \gamma_t \partial [g_t(x_t)]_{+}\hat{q}_{t-1} \rangle \nonumber \\
& + \langle x^{*} - z_{t+1}, \gamma_t \partial [g_t(z_{t+1})]_{+}\hat{q}_{t-1} \rangle.  \nonumber
\end{align} 

Then, expanding the last two terms by using the fact that for any convex function $g(\cdot)$, $\langle y - x, \partial g(x) \rangle \leq g(y)-g(x)$, we get
\begin{align}
& \langle x_t - x^{*}, c_t \rangle \leq \frac{\eta_{t}}{2}\norm{c_t - h_t}^2 \nonumber \\
& + \frac{1}{\eta_t}\left(D_\mathcal{R}(z_t,x^{*}) - D_\mathcal{R}(x^{*},z_{t+1})- D_\mathcal{R}(x_t,z_t)\right) \nonumber \\
& + \gamma_t \langle \hat{q}_{t-1}, [g_t(x^{*})]_{+} \rangle - \gamma_t \langle \hat{q}_{t-1}, [g_t(x_t)]_{+} \rangle.  
\label{eq:pf1-eq4}
\end{align}

Now,
\begin{align}
& \gamma_t \langle \hat{q}_{t-1}, [g_t(x^{*})]_{+} \rangle - \gamma_t \langle \hat{q}_{t-1}, [g_t(x_t)]_{+} \rangle = \gamma_t \langle \hat{q}_{t-1}, [g_t(x^{*})]_{+} \rangle \nonumber \\
& - \gamma_t \langle q_{t-1}, [g_t(x_t)]_{+} \rangle - \gamma^2_t  \langle [g_t(z_t)]_{+},[g_t(x_t)]_{+} \rangle. \nonumber 
\end{align}

Substituting the above in Eq. \eqref{eq:pf1-eq4}, we get
\begin{align}
& \gamma_t \langle q_{t-1}, [g_t(x_t)]_{+} \rangle + \langle x_t - x^{*}, c_t \rangle \leq \frac{\eta_{t}}{2}\norm{c_t - h_t}^2 \nonumber \\
& + \frac{1}{\eta_t}\left(D_\mathcal{R}(z_t,x^{*}) - D_\mathcal{R}(x^{*},z_{t+1})- D_\mathcal{R}(x_t,z_t)\right) \nonumber \\
& + \gamma_t \langle \hat{q}_{t-1}, [g_t(x^{*})]_{+} \rangle - \gamma^2_t  \langle [g_t(z_t)]_{+},[g_t(x_t)]_{+} \rangle.  \label{eq:pf1-eq8} 
\end{align}

Summing over $t \in \{1,...,T\}$, we get
\begin{align}
& \sum_{t=1}^T \gamma_t \langle q_{t-1}, [g_t(x_t)]_{+} \rangle + R_T \leq \sum_{t=1}^T \frac{\eta_{t}}{2}\norm{c_t - h_t}^2 \nonumber \\
& + \sum_{t=1}^T \frac{1}{\eta_t}\left(D_\mathcal{R}(z_t,x^{*}) - D_\mathcal{R}(x^{*},z_{t+1})\right) \nonumber \\
& - \sum_{t=1}^T \frac{1}{\eta_t} D_\mathcal{R}(x_t,z_t) + \sum_{t=1}^T\gamma_t \langle \hat{q}_{t-1}, [g_t(x^{*})]_{+} \rangle\nonumber \\
& - \sum_{t=1}^T\gamma^2_t  \langle [g_t(z_t)]_{+},[g_t(x_t)]_{+} \rangle.  \nonumber 
\end{align}

This proves the first part of the lemma. Next, we observe that
\beq \norm{q_t}_1 = \norm{q_{t-1} + \gamma_t [g_t(x_t)]_{+}}_1 = \norm{q_{t-1}}_1 + \gamma_t \norm{[g_t(x_t)]_{+}}_1. \nonumber \eeq 

Hence,
\begin{align}
& \norm{[g_t(x_t)]_{+}}_1 = \frac{1}{\gamma_t}\left(\norm{q_t}_1 - \norm{q_{t-1}}_1\right) \nonumber \\
& = \frac{1}{\gamma_t}\norm{q_t}_1 -\frac{1}{\gamma_{t-1}}\norm{q_{t-1}}_1 + \left(\frac{1}{\gamma_{t-1}} -\frac{1}{\gamma_t}\right)\norm{q_{t-1}}_1. \nonumber 
\end{align} 

Summing over $t \in \{1,...,T\}$, we get 
\beq \sum_{t=1}^T \norm{[g_t(x_t)]_{+}}_1 = \frac{1}{\gamma_T}\norm{q_T}_1 + \sum_{t=1}^{T-1}\left(\frac{1}{\gamma_t} - \frac{1}{\gamma_{t+1}}\right)\norm{q_t}_1. \label{eq:pf1-eq5} \eeq

Also,
\begin{align} 
& \norm{q_t}^2 = \norm{q_{t-1} + \gamma_t [g_t(x_t)]_{+}}^2 \nonumber \\
& = \norm{q_{t-1}}^2 + 2\gamma_tq_{t-1}[g_t(x_t)]_{+} + \norm{\gamma_t [g_t(x_t)]_{+}}^2. \nonumber
\end{align}

This implies that
\beq \frac{1}{2}\left(\norm{q_t}^2 - \norm{q_{t-1}}^2\right) = \gamma_t \langle q_{t-1}, [g_t(x_t)]_{+} \rangle + \frac{1}{2}\norm{\gamma_t [g_t(x_t)]_{+}}^2. \label{eq:pf1-eq6} \eeq 

Next, we have that
\begin{align}
& \gamma^2_t  \langle [g_t(z_t)]_{+},[g_t(x_t)]_{+} \rangle = \frac{1}{2}\left( \norm{\gamma_t [g_t(z_t)]_{+}}^2 + \norm{\gamma_t [g_t(x_t)]_{+}}^2 \right. \nonumber \\
& \left. - \gamma^2_t\norm{[g_t(z_t)]_{+}-[g_t(x_t)]_{+}}^2 \right). \nonumber \end{align}

Then, using the fact that $g_t(z_t)-g_t(x_t) \leq G \norm{z_t - x_t}$, we get
\begin{align}
& \gamma^2_t \langle [g_t(z_t)]_{+},[g_t(x_t)]_{+} \rangle \geq \frac{\gamma^2_t}{2}\left( \norm{[g_t(z_t)]_{+}}^2 + \norm{[g_t(x_t)]_{+}}^2 \right) \nonumber \\
& - \frac{\gamma^2_t G^2}{2}\norm{z_t-x_t}^2. \label{eq:pf1-eq7} \end{align}

Then, using Equations \eqref{eq:pf1-eq8}, \eqref{eq:pf1-eq5}, \eqref{eq:pf1-eq6}, and \eqref{eq:pf1-eq7}, we get
\begin{align}
& \frac{1}{2}\left(\norm{q_t}^2 - \norm{q_{t-1}}^2\right) + \langle x_t - x^{*}, c_t \rangle \leq \frac{\eta_{t}}{2}\norm{c_t - h_t}^2 \nonumber \\
& + \frac{1}{\eta_t}\left(D_\mathcal{R}(z_t,x^{*}) - D_\mathcal{R}(x^{*},z_{t+1})\right) \nonumber \\
& + \left(\gamma^2_t G^2 - \frac{1}{\eta_t}\right) D_\mathcal{R}(x_t,z_t) + \gamma_t \langle \hat{q}_{t-1}, [g_t(x^{*})]_{+} \rangle. \nonumber
\end{align}

Since $q_0 = 0$, summing over $t \in \{1,...,T\}$, we get
\begin{align}
& \frac{1}{2}\norm{q_T}^2 \leq \sum_{t=1}^T \frac{\eta_{t}}{2}\norm{c_t - h_t}^2  -R_T \nonumber \\
& + \sum_{t=1}^T\frac{1}{\eta_t}\left(D_\mathcal{R}(z_t,x^{*}) - D_\mathcal{R}(x^{*},z_{t+1})\right) \nonumber \\
& + \sum_{t=1}^T\left(\gamma^2_t G^2 - \frac{1}{\eta_t}\right) D_\mathcal{R}(x_t,z_t) + \sum_{t=1}^T \gamma_t \langle \hat{q}_{t-1}, [g_t(x^{*})]_{+} \rangle. \nonumber
\end{align}

This completes the proof $\blacksquare$

\section*{Appendix B: Proof of Theorem \ref{thm:linear-case}}
\label{sec:thm}

From Lemma \ref{lem:intermediate}, we have that
\begin{align} 
& \sum_{t=1}^T \gamma_t \langle q_{t-1}, [g_t(x_t)]_{+} \rangle + R_T \leq \sum_{t=1}^T \frac{\eta_{t}}{2}\norm{c_t - h_t}^2 \nonumber \\
& + \sum_{t=1}^T \frac{1}{\eta_t}\left(D_\mathcal{R}(z_t,x^{*}) - D_\mathcal{R}(x^{*},z_{t+1})\right) \nonumber \\
& - \sum_{t=1}^T \frac{1}{\eta_t} D_\mathcal{R}(x_t,z_t) + \sum_{t=1}^T\gamma_t \langle \hat{q}_{t-1}, [g_t(x^{*})]_{+} \rangle \nonumber \\
& - \sum_{t=1}^T\gamma^2_t  \langle [g_t(z_t)]_{+},[g_t(x^{*})]_{+} \rangle. \nonumber 
\end{align}

Since $x^{*}$ satisfies $g_t(x) \leq 0$ for all $t$, we have that 
\begin{align}
& R_T \leq \sum_{t=1}^T \frac{\eta_{t}}{2}\norm{c_t - h_t}^2 \nonumber \\
& + \sum_{t=1}^T \frac{1}{\eta_t}\left(D_\mathcal{R}(z_t,x^{*}) - D_\mathcal{R}(x^{*},z_{t+1})\right). \nonumber 
\end{align}

Therefore, since $\eta_t = \eta$ is a constant, we have that
\beq
R_T \leq \frac{\eta}{2} \sum_{t=1}^T \norm{c_t - h_t}^2 + \frac{1}{\eta} D_\mathcal{R}(z_1,x^{*}). \nonumber 
\eeq

This completes the first part of the proof. Once again, from Lemma \ref{lem:intermediate}, we have that
\begin{align}
& \frac{1}{2}\norm{q_T}^2 \leq \sum_{t=1}^T \frac{\eta_{t}}{2}\norm{c_t - h_t}^2  -R_T \nonumber \\
& + \sum_{t=1}^T\frac{1}{\eta_t}\left(D_\mathcal{R}(z_t,x^{*}) - D_\mathcal{R}(x^{*},z_{t+1})\right) \nonumber \\
& + \sum_{t=1}^T\left(\gamma^2_t G^2 - \frac{1}{\eta_t}\right) D_\mathcal{R}(x_t,z_t) + \sum_{t=1}^T \gamma_t \langle \hat{q}_{t-1}, [g_t(x^{*})]_{+} \rangle. \nonumber
\end{align}

Since $\left(\gamma^2_t G^2 - \frac{1}{\eta_t}\right) = 0$ and $\eta_t = \eta$, we get that
\beq
\frac{1}{2}\norm{q_T}^2 \leq \frac{\eta}{2} \sum_{t=1}^T \norm{c_t - h_t}^2 + \frac{1}{\eta} D_\mathcal{R}(z_1,x^{*}) -R_T. \nonumber
\eeq

Since $-R_T \leq FT$, we get that
\beq
\frac{1}{2}\norm{q_T}^2 \leq \frac{\eta}{2} \sum_{t=1}^T \norm{c_t - h_t}^2 + \frac{1}{\eta} D_\mathcal{R}(z_1,x^{*}) + FT. \nonumber
\eeq

Since
\beq \sum_{t=1}^T \norm{[g_t(x_t)]_{+}}_1 = \frac{1}{\gamma_T}\norm{q_T}_1 + \sum_{t=1}^{T-1}\left(\frac{1}{\gamma_t} - \frac{1}{\gamma_{t+1}}\right)\norm{q_t}_1, \nonumber \eeq
and $\gamma_t = \gamma$, we get that
\begin{align}
    & \sum_{t=1}^T \norm{[g_t(x_t)]_{+}}_1 \nonumber \\
    & \leq  \frac{1}{\gamma}\sqrt{\frac{\eta}{2} \sum_{t=1}^T \norm{c_t - h_t}^2 + \frac{1}{\eta} D_\mathcal{R}(z_1,x^{*}) + FT} \blacksquare \nonumber
\end{align} 


\end{document}